\documentclass[journal]{IEEEtran}
\usepackage{cite}

\ifCLASSINFOpdf
  \usepackage[pdftex]{graphicx}
\else
\fi
\usepackage{amsmath}
\usepackage{algorithmic}

\usepackage{array}
\usepackage{multirow}
\begin{document}
%
\title{A Real-time Robotic Grasp Approach with Oriented Anchor Box}
%
%
%

\author{Hanbo~Zhang, Xinwen~Zhou,
	Xuguang~Lan,~\IEEEmembership{Member,~IEEE,}
	Jin~Li, Zhiqiang~Tian,
        and~Nanning~Zheng~\IEEEmembership{Fellow,~IEEE}
\thanks{*This work was supported in part by the key project of Trico-Robot plan of NSFC under grant No. 91748208, National Key Program of China No.2017YFB1302200,key project of Shaanxi province No.2018ZDCXL-GY-06-07, and NSFC No.61573268.}
\thanks{Hanbo Zhang and Xuguang Lan are with the Institute of Artificial Intelligence and Robotics, the National Engineering Laboratory for Visual Information Processing and Applications, School of Electronic and Information Engineering,
        Xi'an Jiaotong University, No.28 Xianning Road, Xi'an, Shaanxi, China.
        {\tt\small zhanghanbo163@stu.xjtu.edu.cn, xglan@mail.xjtu.edu.cn}}}
\maketitle

\begin{abstract}
Grasp is an essential skill for robots to interact with humans and the environment. In this paper, we build a vision-based, robust and real-time robotic grasp approach with fully convolutional neural network. The main component of our approach is a grasp detection network with oriented anchor boxes as detection priors. Because the orientation of detected grasps is significant, which determines the rotation angle configuration of the gripper, we propose the Orientation Anchor Box Mechanism to regress grasp angle based on predefined assumption instead of classification or regression without any priors. With oriented anchor boxes, the grasps can be predicted more accurately and efficiently. Besides, to accelerate the network training and further improve the performance of angle regression, Angle Matching is proposed during training instead of Jaccard Index Matching. Five-fold cross validation results demonstrate that our proposed algorithm achieves an accuracy of 98.8\% and 97.8\% in image-wise split and object-wise split respectively, and the speed of our detection algorithm is 67 FPS with GTX 1080Ti, outperforming all the current state-of-the-art grasp detection algorithms on Cornell Dataset both in speed and accuracy. Robotic experiments demonstrate the robustness and generalization ability in unseen objects and real-world environment, with the average success rate of 90.0\% and 84.2\% of familiar things and unseen things respectively on Baxter robot platform.
\end{abstract}


%
\IEEEpeerreviewmaketitle

\section{Introduction}
%
%
%
%
\IEEEPARstart{R}{eliable} grasp is the base of robot-environment interaction and human-robot cooperation. The advance in grasp can boost the widespread use of robots in society to promote the development of productivity. Human beings can accomplish grasp under different circumstances almost instinctively. However, for robots, grasp is still a challenging task due to that grasp needs combination of perception, planning and execution. To grasp an object successfully, robots need first locate targets and predict grasps for this object. A precise detection of grasps makes way for subsequent operations.

\begin{figure}[t] 
 \center{\includegraphics[scale=0.3]{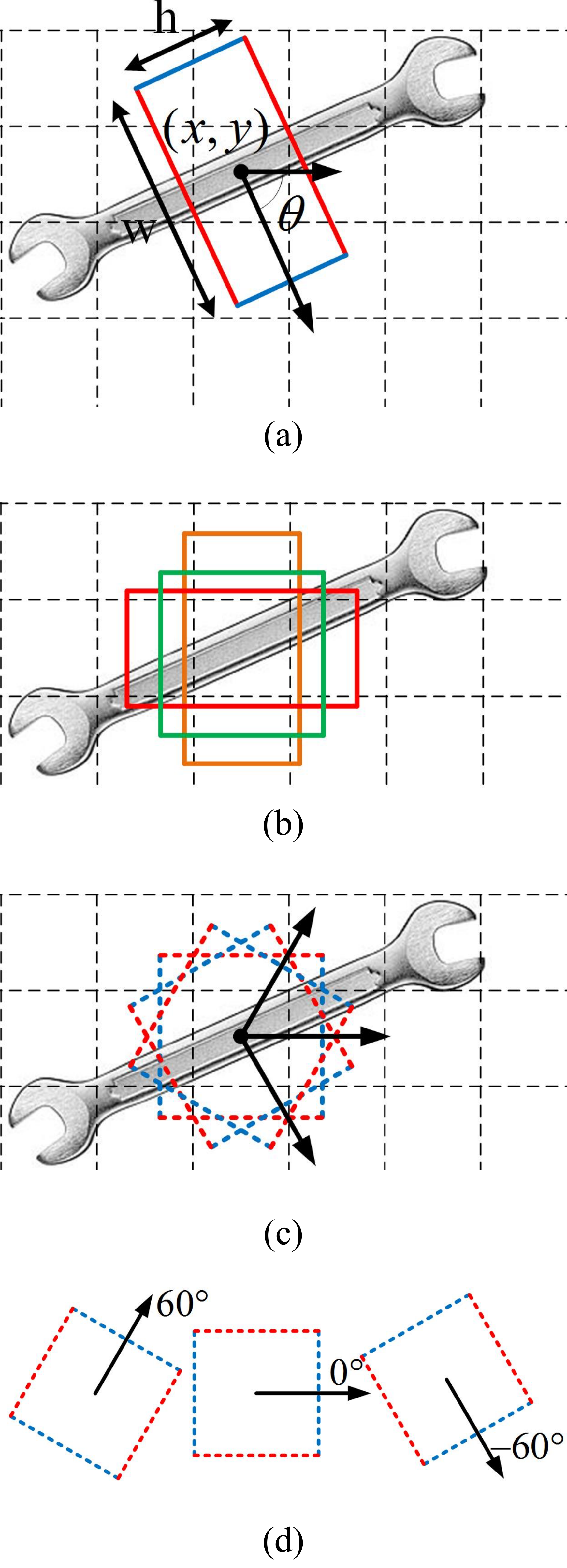}}        
 \caption{(a) An example of five-dimensional representation of the grasp for a wrench. (b) Non-oriented reference rectangles with 1 scale and 3 aspect ratios. Aspect ratio for red, green and orange rectangle is 1:2, 1:1 and 2:1 respectively. (c) and (d) Our oriented anchor boxes in a grid cell. The orientations of these oriented anchor boxes is $60^{\circ}$, $ 0^{\circ}$ and $-60^{\circ}$.}
 \label{intro}
 \end{figure}

Recent works have explored the performance of deep learning for robotic grasp detection\cite{guo2016grasp,chu2018grasp}. They usually take RGB or RGB-D images as input to do regression on grasp rectangles and achieve better performance than previous works. As shown in object detection\cite{ssd,fasterrcnn}, the prior information plays an important role in the final performance since it reduce the difficulty of direct regression of coordinates with regression of offsets. Inspired by these works, Guo et al. \cite{guo2017hybrid} introduce reference rectangles, also known as anchor boxes without varying orientation, in grasp detection shown in Fig. \ref{intro}(b). Reference rectangles are a set of rectangles overlaid on the image at different spatial locations. In Fig. \ref{intro}(b), a set of reference rectangles with 1 scale and 3 aspect ratios are at a spatial location. These rectangles are parallel to horizontal axis. The orientation of the grasp rectangles is quantized and predicted by classification. 

The orientation matters much more to robotic grasp detection \cite{lenzgrasp}. In most cases, a feasible gripper orientation for a given location is limited to a small range and is closely relevant to the location. Inspired by this, our proposed algorithm applies Oriented Anchor Box Mechanism as grasp priors in deep learning grasp detection network to predict grasps from processed images.

Differently from the previous works, as shown in Fig.\ref{intro}(c), each pixel in feature map associates with an area (defined as a cell) on the input image. For each cell we set several oriented prior rectangles with different default rotation angles as the reference for the final grasp detection, which is defined as Oriented Anchor Box Mechanism in this paper. Fig. \ref{intro}(c) and (d) demonstrates an example of the proposed oriented anchor boxes centered at one grid cell with the same scale and three default values ($60^{\circ}$, $0^{\circ}$ and $-60^{\circ}$). In our work, the grasps are represented as oriented rectangle, which has five parameters: $(x, y, w, h, \theta)$ as shown in Fig. \ref{intro}(a). This is different from the horizontal anchor box in objection detection. The introduction of angle term is from the consideration of the property of grasp detection. The target of our algorithm is to refine and classify these oriented anchor boxes: the locations are refined by regression layer and a proposed classifier is used to estimate whether these oriented anchor boxes are graspable or not.

Our model consists of two parts, a feature extractor backbone and a predictor, which includes a regression layer and classification layer. Our feature extractor backbone takes an RGB or RGD image as input and produce feature maps for grasp detection. To combine depth modalities with RGB information, the Blue channel in the original RGB image is substituted by normalized depth information because depth information can help the robot sense spatial structures of the environment and improve performance of our models better than B channel.

Besides, during training stage, a new matching strategy named Angle Matching is proposed instead of the old matching strategy based on Jaccard index (.i.e., the intersection over union (IoU) between two slant rectangles). The motivation behind this change is as follow: 1) Computation of angle difference is much less time-consuming than Jaccard index. Therefore, it will result in a faster training than before. 2) The fact that if the center of the ground-truth rectangle is inside of a grid cell, at least half of its surface is overlap with the grid cell guarantees a large overlap between ground-truth and its corresponding oriented anchor box no matter what the rotation angle is. These indicate that using Angle Matching is a smarter way to train the grasp detection network. Details will be discussed in the following sections.


In robotic grasp experiment, we use grasp detection model to find the grasp point and grasp vector. The grasp point is defined as the closest point away from the RGB-D camera and the grasp vector is the average of surface normals near the grasp point. Using coordinate calibration, we map the grasp vector from camera coordinates to Baxter coordinates, and make Baxter robot execute the grasp configuration using inverse kinematics. 

In summary, our contributions in this paper are as follow:
\begin{itemize}
\item In this paper, we propose a grasp detection algorithm based on fully convolutional neural network, which is 2.8\% and 1.5\% higher in image-wise split and object-wise split respectively in accuracy and 7 times faster in speed than the two-stage grasp detection algorithm of Chu et al\cite{chu2018grasp}. 
\item We propose the Oriented Anchor Box Mechanism for robotic grasp detection considering the property of robotic grasps and its improvement on grasp detection tasks is proved in our experiments. Moreover, benefitting from Oriented Anchor Box Mechanism, our method can directly regress angles of predicted grasps with oriented priors instead of classification.
\item A new matching strategy called Angle Matching is proposed in this paper to accelerate network training without loss of performance, which help us save about 10\% of training time.
\end{itemize}

The rest of this paper is organized as follow: Section II reviews the background and related works of our approach; Section III describes the details of our grasp detection algorithm; Section IV gives the experimental results and discussions on Cornell Grasp Dataset; and finally, conclusions of this paper are discussed in Section V.
 
\section{Related Work}

As a part of robotics, robotic grasp has been studied for a long time\cite{graspsurvey}. At the early stage, Kaneko et al.\cite{kaneko2000scale} study the robotic grasp pattern inspired by grasping action of humans. Simple models are used to test the grasp performance. Later, robotic grasp methods are developed based on 3D models\cite{graspit, svmgrasp, liu2014vision}. In \cite{graspit} and \cite{svmgrasp}, by adjusting rendering condition, 3D object datasets used to train and test the proposed methods are larger than the dataset collected by human. However, full 3D models are often not known and difficult to obtain in the real scene. Therefore, researchers consider to combine multi-model information to help robots plan grasp, such as visual and tactile sensing\cite{sun2016grasp,guo2017hybrid} and RGB-D information\cite{jiang2011grasp,lenzgrasp}.

\begin{figure*}[t] 
 \center{\includegraphics[width=0.9\textwidth]{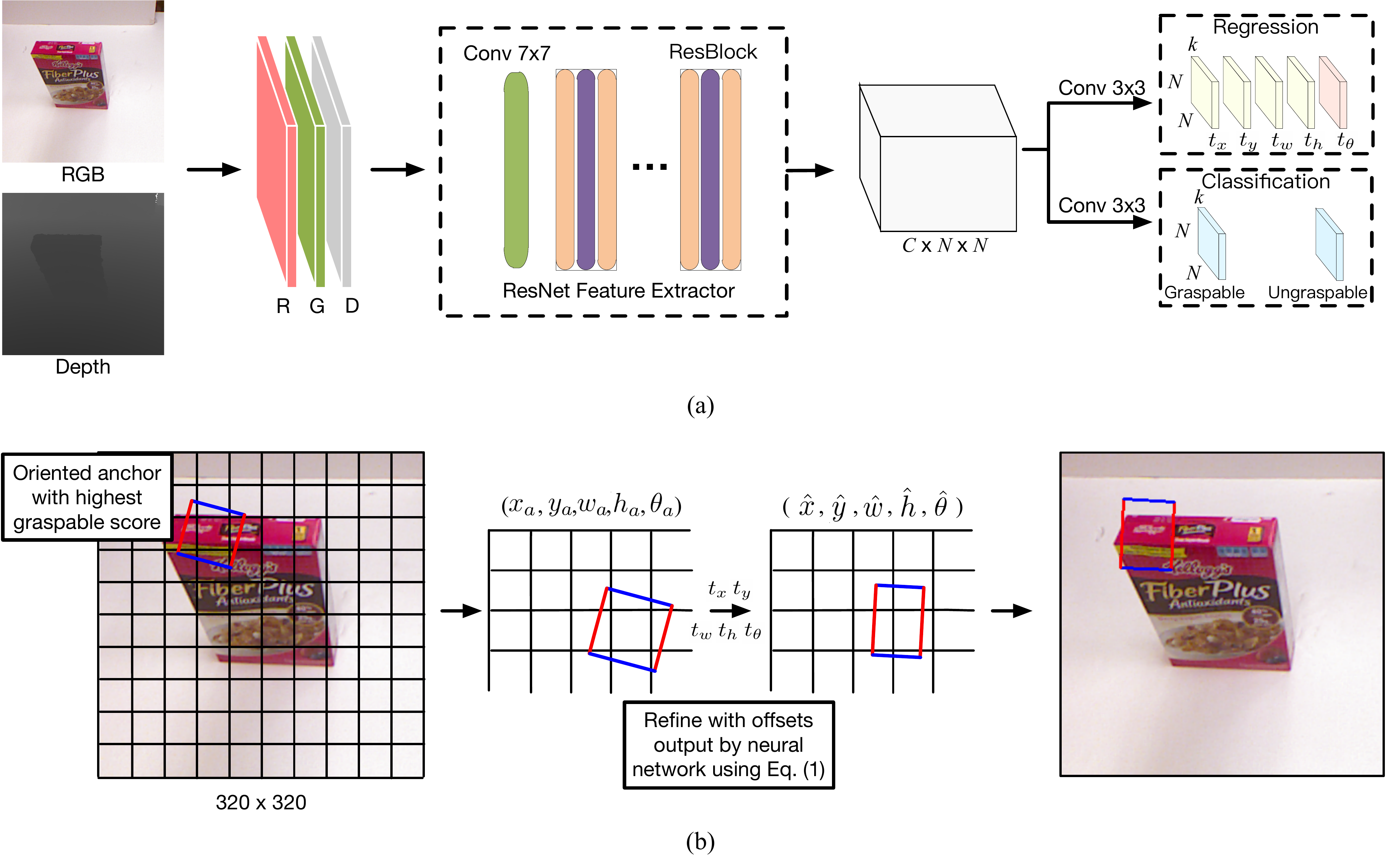}}        
 \caption{(a) Network architecture based on ResNet-Conv5. The input is a $320\times320$ RGD image. The output includes regression and classification results. Note that size of feature maps used to predict the final results depends on the backbone and feature layer (e.g. if we use VGG16 as backbone and Conv4 features, the feature size will be $512\times10\times10$). (b) Process using network output to compute grasp prediction. First, the oriented anchor box with highest graspable score is found according to classification results. Then, grasp prediction is computed using Eq. \ref{eq1}} 
 \label{fig2}
 \end{figure*}
 
Deep learning and popularization of consumer depth cameras make it possible to use RGB-D images to predict high quality grasps. Features extracted through deep learning is proved to be better than hand engineered ones\cite{dl}. This improvement has been explored in many fields such as classification\cite{imagenet}, object detection\cite{ssd,fasterrcnn}, action recognition\cite{actionrecogTSMC} and pose estimation\cite{liang2018limb}. Especially, the improvement of object detection has inspired many researchers in the field of grasp detection.

Lenz et al.\cite{lenzgrasp} first employ neural network as classifier to predict grasps from image patches, which are cropped from original image using sliding window. Furthermore, they present a five-dimension configuration of robotic grasp. It can be projected to 3D grasp vector. Therefore, methods of object detection are possible to be transferred to grasp detection. Their algorithm reaches an accuracy of 73.9\% accuracy on image-wise split while running at a speed of 13.5 seconds per frame. It is caused by time-consuming sliding window method. This algorithm can not be used in real-time systems.

Redmon et al.\cite{redmongrasp} speed up grasp detection by dividing the image into $N \times N$ grid cells and regressing a grasp rectangle in each grid cell with the probability that the predicted grasp would be feasible. By doing this, they pass the entire image only once to predict grasp rectangle instead of passing several patches. This method also enables the simultaneous prediction of several grasp rectangles. Meanwhile, the introduction of depth information also contributes to the improvement of accuracy. A relatively shallow network, AlexNet\cite{imagenet}, is used to extract features, because deeper networks seem to be prone to degenerate at that time. This algorithm, in turn, inspires researchers in object detection. With the appearance of the more reliable and deeper network, ResNet, the algorithm of Kumra et al.\cite{resnetgrasp} directly regresses a grasp rectangle from input image. By introducing a deeper network, the accuracy of Direct Regression model is 88.84\%, which outperforms the work of Redmon et al.\cite{redmongrasp} by 4.13\%.

Like Redmon et al, the algorithm of Guo et al.\cite{guo2016grasp} regress grasp rectangles for each grid cell. Instead of only one grasp prediction in each grid cell, they associate each grid cell with several horizontal reference rectangles with multiple scales and aspects. Here, rotation angle is seperated from location regression terms and is predicted by classification. Each reference rectangle will have its graspable and ungraspable scores. Guo et al.\cite{guo2017hybrid} go further to collect THU Grasp Dataset which contains tactile information and integrate it in their model. On image-wise split of Cornell Grasp Dataset, they reach an accuracy of 93.2\% with only visual information.
 
The classification of graspable or ungraspable is then combined with angle classification as a single classification problem by Chu et al\cite{chu2018grasp}. The classification problem contains 20 labels, which includes 19 rotation angles and No Orientation competing class. Under this framework, algorithms in object detection can be transferred to grasp detection with little effort. Chu et al reset oriented ground truth grasp rectangle to have vertical weight and horizontal height. Thus, Faster-RCNN\cite{fasterrcnn} can be applied to grasp detection. Their model reach the state-of-the-art perform with the accuracy of 96.0\% on image-wise split.

In this paper, we go further in robotic grasp detection with fully convolutional neural network and oriented anchor boxes, which are proved to be more efficient and suitable for location of grasp than vertical ones. Moreover, we propose a new matching strategy called Angle Matching instead of matching based on Jaccard index. It is proved to provide the network a much shorter training time. Finally, benefitting from Oriented Anchor Box Mechanism, our method can directly regress angles of predicted grasps based on oriented anchor boxes instead of classification.

\section{Fully Convolutional Grasp Detection}

\subsection{Network Architecture}
 
  \begin{figure}[t] 
 \center{\includegraphics[scale=0.4]{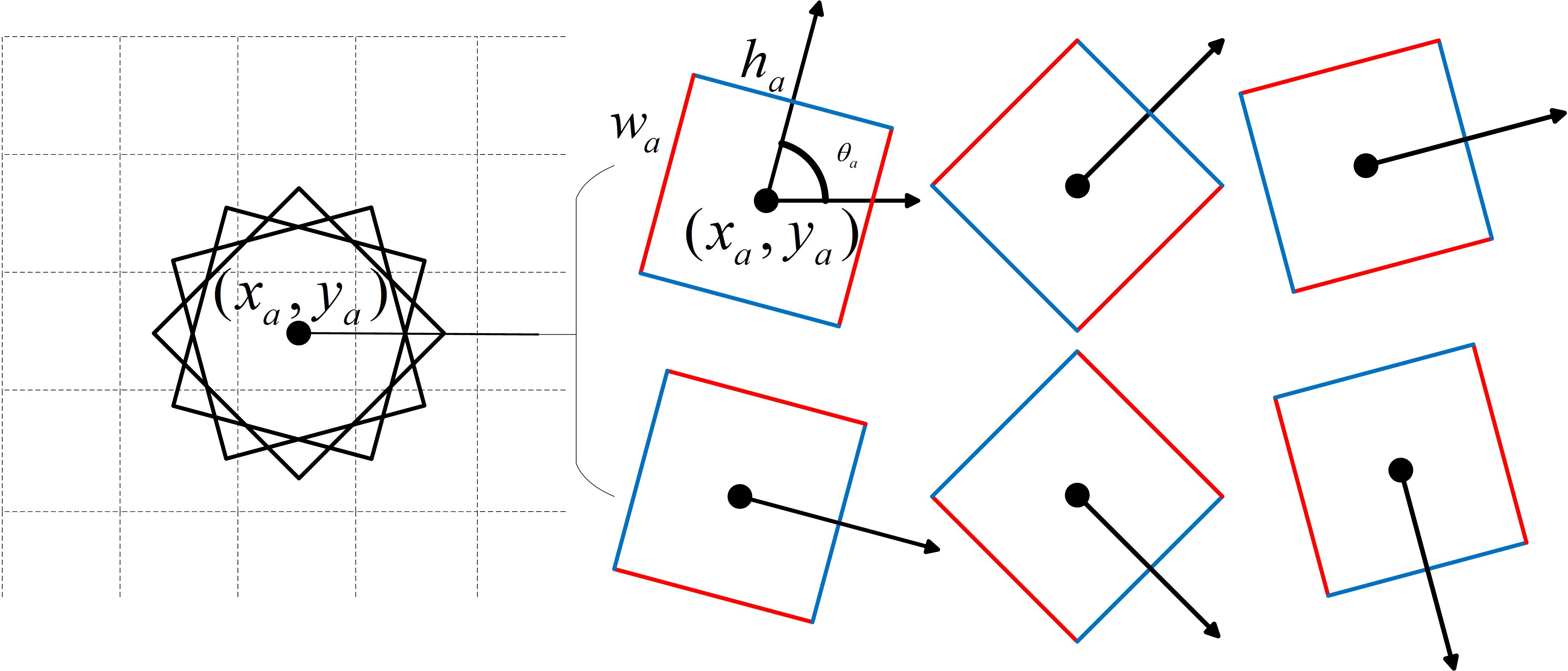}}        
 \caption{Oriented Anchor Box Mechanism. The origin input image is devided into grid cells according to the feature maps output by the backbone. In each cell, there are $k$ predefined oriented anchor boxes to provide prior information for the final prediction. The only one scale of anchors depends on size of grid cell and the aspect ratio is always 1:1. Any two adjacent oriented anchor box have a $180/k^{\circ}$ rotation angle difference.}
 \label{fig3}
 \end{figure}
 
As most detection algorithms, our model consists of two parts: a backbone for feature extraction and a header for classification and regression. During experiments, we try different backbones in our network including VGG16\cite{vgg}, ResNet50 and ResNet101\cite{resnet}. They are all pretrained on ImageNet.

For all the backbones, we use the feature maps output by the last layer of the desired stage. Note that size of feature maps $C\times N\times N$ used to predict the final results depends on the backbone and feature layer. $C$ is the channel number ($e.g.$ $1024$ for ResNet layer3). Each pixel in the feature map of size $N \times N$ corresponds to a cell in the original image. Formally, we use C4 and C5 denote the output of conv4 and conv5 in ResNets or VGG16, which have strides of 16 and 32 pixels with respect to the original image, respectively. The input image size of our network is $320 \times 320$, and the feature map used to predict grasps is $20\times20$ for C4 and $10\times 10$ for C5. According to feature map size, the input image is also divided into $N\times N$ grid cells and the scale of oriented anchor box is set according to the cell size. Our network architecture based on ResNet is shown in Fig.\ref{fig2}(a).
 
For the detector, we have a convolutional layer of kernel size $3\times 3$ for classification, which outputs $2\times k$ channels with size $N\times N$. Each 2 channels represents the graspable and ungraspable scores of all the oriented anchor boxes with the same rotation angle. The visualization of these feature map will be shown in experiments section. We also use a convolutional layer to produce $5\times k$ offset values for each location in feature map. Our model is a fully convolutional network, which has much less parameters compared with those models that have fully connected layer. Process that uses output of the network to obtain the final prediction results is shown in Fig. \ref{fig2}(b). First, the oriented anchor box with highest graspable score is found according to classification results. Then, grasp prediction is computed using Eq. \ref{eq1} from oriented anchor box and offsets output by the network.

\subsection{Oriented Anchor Box Mechanism}

The anchor box mechanism which is proposed in object detection and proved to vastly improve the performance of object detection algorithms inspires us to do similar work in robotic grasp detection. However, different from the vertical ground truth bounding boxes in object detection, grasp rectangles are always slant since the object can be put arbitrarily. Therefore, we propose Oriented Anchor Box Mechanism in our work. The predefined oriented prior rectangles are called oriented anchor boxes in this paper, instead of the vertical ones, and corresponding to this change, one more dimension need to be added into anchor representation. In detail, the oriented anchor box includes 5 dimensions: $(x_a,y_a,w_a,h_a,\theta_a)$. $(x_a,y_a)$ represents the center coordinate, $w_a$ and $h_a$ are the width and height and $\theta_a$ is the counterclockwise rotation angle of the oriented anchor box from horizon. All oriented anchor boxes provide the grasp prediction network prior information for regression, and make it possible to predict the grasp rectangle offsets relative to oriented anchor boxes including not only the center coordinate, width and height, but also the angle, which is much easier than directly regression or classification.

For each picture, the output of the feature extractor backbone is a $C \times N \times N$ matrix. Therefore, the input image is divided into $N \times N$ grid cells and each grid cell is associate with $k$ oriented anchor boxes. Oriented anchor boxes of a grid cell have the same center coordinate and size ($48\times48$ for stride 32 and $24\times24$ for stride 16). These boxes are different with each other in rotation angle. The location regression layer, which is in fact a convolutional layer with kernel size $3\times3$, slides over the feature map and outputs $5\times k$ values for each pixel of the feature maps associating with a grid cell in the origin image. In other words, five terms $(t_{\hat{x}},t_{\hat{y}},t_{\hat{w}},t_{\hat{h}},t_{\hat{\theta}})$ are regressed in this layer for each oriented anchor box. These five terms are the offsets of the predefined oriented anchor box configuration. Following Eq. \ref{eq1}, we will get the final predicted grasp rectangle.

\begin{equation}
\begin{split}
t_{\hat{x}}&=(\hat{x}-x_{a})/w_{a}\\
t_{\hat{y}}&=(\hat{y}-y_{a})/h_{a}\\
t_{\hat{w}}&=log(\hat{w}/w_a)\\
t_{\hat{h}}&=log(\hat{h}/h_a)\\
t_{\hat{\theta}}&=(\hat{\theta}-\theta_{a})/(90/k)\end{split}
\label{eq1}
\end{equation}

In Eq. \ref{eq1}, $t$ is a five dimension offset vector representing the output of regression layer of our network, and $(\hat{x},\hat{y},\hat{w},\hat{h},\hat{\theta})$ is the final prediction of grasp rectangles.

Unlike object detection dataset, the size of grasp rectangle ground truth labeled in the Cornell Grasp Dataset varies slightly. As shown in Fig.\ref{fig3}, our oriented anchor boxes only have one default scale and its default aspect ratio is always 1:1. Any two adjacent oriented anchor boxes have a $180/k^{\circ}$ rotation angle difference. The size of feature map is also a very important parameter. A larger size can bring a denser prediction. Therefore, in experiments, we try different scale and feature settings to see its influence on the final performance.

\subsection{Angle Matching}

Previous works use Jaccard index to distinguish positive and negative examples during training. However, computation of Jaccard index between two slant rectangles is much more complete than two vertical ones, which is very time-consuming and can not be computed by batch in practice.

To improve this and accelerate training of network, in our work, a new matching strategy is proposed based on angle difference between ground truth grasp rectangles and oriented anchor boxes called Angle Matching. In detail, we assign positive label to oriented anchor box that satisfied following two conditions:

\begin{itemize}
\item 1) The centers of both the ground-truth rectangle and the oriented anchor box should be located in the same grid cell. 
\item 2) The difference between rotation angles of ground-truth rectangle and matched oriented anchor box should be less than $90^{\circ}/k$ ($k$ is the anchor number in one cell).
\end{itemize}

\begin{table}[b]
\caption{Comparison between Jaccard Index Matching (JM) and Angle Matching (AM).}
\label{angmatch}
\begin{center}
\begin{tabular}{cccccc}

\hline
\multirow{2}{*}{\bf{Anc.}}&\multirow{2}{*}{\bf{Feat.}} & \multicolumn{4}{c}{\bf {JM / AM}} \\
\cline{3-6}
 & & $ \overline{\Delta Ang}$& $\overline{IoU}$ &$t_{match}$(ms)& $t_{iter}$(s)\\
\hline
\multirow{2}{*}{k=4}	& C4 & 60 / 12 & 0.58 / 0.35 & 40.40 / 0.07 & 0.49 / 0.45 \\
& C5 & 60 / 12 & 0.56 / 0.35 & 37.74 / 0.07 & 0.52 / 0.49 \\
\multirow{2}{*}{k=6}	 & C4 & 62 / 8 & 0.57 / 0.35 & 59.45 / 0.07 & 0.51 / 0.46  \\
& C5 & 62 / 8 & 0.56 / 0.36 & 56.49 / 0.07 & 0.55 / 0.49  \\
\multirow{2}{*}{k=8}	 & C4 & 61 / 6 & 0.58 / 0.35 & 81.61 / 0.07 & 0.54 / 0.45\\
& C5 &60 / 6 & 0.56 / 0.35 & 76.38 / 0.07 & 0.56 / 0.49 \\
\hline

\end{tabular}
\end{center}
\end{table}

Grasp detection should consider rotation angle as well as localization. Under first condition, the center of the matched oriented anchor box is the nearest one to the center of ground-truth grasp rectangle. The second condition ensures that the matched oriented anchor box and ground-truth grasp rectangle are within the same range of rotation angle. 


Obviously, matching using angle difference is much easier than computing the Jaccard index while taking rotation angle into consideration. Previous works assign a reference rectangle (horizontal parallel rectangle) to be positive if its Jaccard index with ground-truth rectangle is above 0.5, which neglects the importance of angle and more time-consuming. 

The comparison between Jaccard Index Matching (JM, left) and Angle Matching (AM, right) is shown in Table \ref{angmatch}, where $ \overline{\Delta Ang}$ and $\overline{IoU}$ represent the average angle difference and Jaccard Index between oriented anchor box and the matched ground truth respectively. $t_{match}$ is the matching time spent in each iteration and $t_{iter}$ is the total iteration time including forward and backward propagation. 

As shown in Table \ref{angmatch}, our matching strategy takes both location and rotation angle into account at a very low computation cost. Though the average IoU of Angle Matching is lower than that of Jaccard Index Matching. However, different from object detection, grasp orientation configuration is more important to help adjust the gripper to fit grasps, which comes from detected rotation angles.  Therefore, Angle Matching is enough to get good matching results without computation of IoU between ground truth and oriented anchor box, and will save about 10\% of the training time. From Table \ref{angmatch} we can see that the more anchor boxes are used, the more time will be saved by Angle Matching. Besides, Angle Matching limits the rotation angle difference between oriented anchor box and ground truth grasp rectangle with accepted loss of IoU. It makes the regression of grasp rotation angles easier and accelerate the convergence of regression loss. Jaccard Index Matching will bring a much larger rotation angle difference than Angle Matching, which makes the regression of grasp rotation angles difficult.

To be fair, in experiments, we use Jaccard index for evaluation, as \cite{guo2017hybrid,redmongrasp,resnetgrasp,chu2018grasp} do. More detailed evaluation will be discussed in Section IV-D.

\subsection{Loss Function}

Loss function of our network includes two parts: classification loss $L_{cls}$ and regression loss $L_{reg}$. 

In grasp rectangle regression, our network outputs five dimension offset for each oriented anchor box as shown in Eq. \ref{eq1}. Therefore, the ground truth needs to be encoded in the same way:

\begin{equation}
\begin{split}
t_{x}&=(x-x_{a})/w_{a}\\
t_{y}&=(y-y_{a})/h_{a}\\
t_{w}&=log(w/w_a)\\
t_{h}&=log(h/h_a)\\
t_{\theta}&=(\theta-\theta_{a})/(90/k)\end{split}
\label{eq2}
\end{equation}

In Eq. \ref{eq2}, each ground truth is mapped into offset space with assigning each ground one or more oriented anchor boxes. Matching strategy used here is Angle Matching introduced in section III.C. The anchor boxes that match at least one ground truth grasp rectangle are also assigned positive labels. Therefore, the loss of regression can be written as Eq. \ref{lreg}:

\begin{equation}
L_{reg}(t)=\sum_{i\in Positive}^P \sum_{m\in \{x,y,w,h,\theta\}}^{}smoothL1(t_m^{(i)}-\hat{t}_m^{(i)})
\label{lreg}
\end{equation}

where we use $Smooth$ $L1$ loss to optimize the regression results. $P$ is the number of positive oriented anchor boxes. $t_{m}$ is the offset predicted by the network. $\hat{t}_{m}$ is the corresponding ground-truth offset value. $t^{(i)}$ is a vector representing the 5 parametrized grasp offset values of the matched oriented anchor box. Parameterization is shown in Equation(2). $\hat{t}^{(i)}$ is the regression result of positive samples from network prediction selected by Angle Matching.

Intuitively, the regression layer gives an offset that helps the corresponding oriented anchor box get more close to the matched ground truth grasp rectangle in position, size and rotation angle. The closer the prediction is with the matched ground truth grasp rectangle, the lower $L_{reg}$ becomes.

The classification loss $L_{cls}$ is defined as cross entropy loss of positive and negative samples. Positive oriented anchor boxes are only small part of the entire oriented anchor boxes, and the rest are negative. Due to this imbalance between positive and negative examples, $L_{cls}$ will lead to no convergence by back propagating. Therefore, we only back-propagate the classification loss of positive examples and part of negative examples. Specifically, the number of negative examples used to optimize classification is three times of the number of positive examples. We sort the negative oriented anchor boxes by their graspability confidence and select the top $3N$ boxes as negative examples used in training, which is called Hard Mining\cite{ssd}. $L_{cls}$ is defined as:

\begin{equation}
L_{cls}(p)=\sum_{i\in Positive}^P log(p^{(i)}_{g}) + \sum_{i\in Negative}^{3P} log(p^{(i)}_{u})
\end{equation}

where $p^{(i)}_{g}$ is the graspability confidence score for the positive sample, $p^{(i)}_{u}$ is the ungraspability confidence score for the negative sample, $p^{(i)}$ is combination of $p^{(i)}_{g}$ and $p^{(i)}_{u}$.

Finally, our loss function is defined as:

\begin{equation}
L(p,t)=\frac{1}{4P}(L_{cls}(p)+\alpha L_{reg}(t))
\end{equation}

The classification loss and regression loss are normalized with number of training samples $4P$ and balanced weight $\alpha$. In our work, $\alpha$ is set to 2.

\section{Validation}

\subsection{Dataset}

In order to compare with other algorithms, our models are trained and tested on Cornell Grasp Dataset. The dataset contains 885 images of 240 graspable objects. In each image, legal grasps are labeled as positive grasp rectangles. For a given object, its grasp rectangles are varied in location, scale and orientation.

Like previous works, we divide the dataset into five folds following 2 ways:
\begin{itemize}
\item 1) {\bf Image-wise split} divides the images into training set and test set at random. This aims to test the generalization ability of the network to new position and orientation of an object it has seen before.
\item 2) {\bf Object-wise split} divides the dataset at object instance level. All the images of an instance are put into the same set, which is to test the generalization ability of the network to new object.
\end{itemize}
To evaluate the performance of our network, we do 5-fold cross validation in each experiment and use the average accuracy as the final results.

The training process for a deep neural network needs a large manually labeled dataset. While this kind of dataset is unavailable in most robotics applications. Researchers solve this problem from two aspects. First, pretrain the network in larger dataset\cite{transfer} like ImageNet\cite{deng2009imagenet}. Second, expand the target dataset by data augmentation. In our work, we apply both techniques in training.

Compared with other datasets in deep learning, the Cornell Grasp Dataset is a small dataset. Therefore, extensive data augmentation is needed before feeding the data into the network. The data augmentation expands the dataset from different aspects. We take a center crop of 320x320 pixels with randomly translation up to 50 pixels in both x and y directions. This image patch is then randomly rotated up to 30 degrees in both clockwise and anti-clockwise direction. Then the image is randomly flipped horizontally or vertically. Then we put the image into the network at the resolution of 320x320. Our augmentation is implemented online, which means every input image is a new image from pixel-level.

\subsection{Implementation Details}

The feature extractors are convolutional layers pretrained on RGB images of ImageNet, which helps the large convolutional neural network to avoid overfitting, especially when the dataset is limited. In experiments, we use the output of C4 and C5 as the features with strides of 16 and 32 pixels with respect to the original image, respectively.

Our models is implemented with Pytorch for its flexibility. For training and testing, our models run on a single NVIDIA GTX1080Ti (Pascal Architecture) with 11GB memory. For each of the models we tested, we employ the same training regimen. The batch size is set as 16. Each model is trained end-to-end for 1000 epochs. We use SGD with momentum of 0.9 to optimize our models. The learning rate is set as 0.002 for RGD input and 0.001 for RGB input with a learning rate decay of 0.0002 and 0.0001, respectively, since the RGD input does not fit the RGB-pretrained model as well as the RGB input.

\subsection{Evaluation Metrics}

Similar to \cite{guo2017hybrid,redmongrasp,resnetgrasp,chu2018grasp}, we also use the rectangle metric to evaluate grasp detection results. In this metric, a predicted grasp is regarded as a good grasp if it satisfies both:
\begin{itemize}
\item 1) The rotation angle difference between the predicted grasp and the ground-truth grasp is within 30�.
\item 2) The Jaccard index of the ground-truth grasp and the predicted grasp is larger than 25\%.
\end{itemize}
The Jaccard index is defined as:
\begin{equation}
J(G,\hat{G}) = \frac{G\cap{\hat{G}}}{G\cup{\hat{G}}}
\end{equation}
where $G$ is the area of the predicted grasp rectangle and $\hat{G}$ is the area of the ground-truth grasp rectangle. $G\cap{\hat{G}}$ is the intersection of these two rectangles. $G\cup{\hat{G}}$ is union of these two rectangles. Note that Jaccard index is only used in our evaluation, not in the matching strategy.

\subsection{Results}

\begin{table}[ht]
\caption{Self-Comparison Experiments Under Different Input Data and Features.}
\label{rgbrgd}
\begin{center}
\begin{tabular}{l|l|c|c}

\hline
\bf{Backbone} & \bf{Setting} & \bf{Image-wise(\%) }& \bf{Object-wise(\%)}\\
\hline
\multirow{4}{*}{\bf{VGG-16}}	& RGB, C5, k=6 & 96.4 & 95.6\\
& RGD, C5, k=6 & 96.6 & 95.5\\
& RGD, C4, k=6 & {\bf98.2} & 96.2\\
& RGD, C4, k=4 & 97.5 & {\bf96.4}\\
\hline
\multirow{4}{*}{\bf{ResNet-50}}	& RGB, C5, k=6 & 98.0 & 95.6\\
& RGD, C5, k=6 & 97.9 & 96.9\\
& RGD, C4, k=6 & 98.3 & 96.7\\
& RGD, C4, k=4 & \bf{98.8} & \bf{97.0}\\
\hline
\multirow{4}{*}{\bf{ResNet-101}}	& RGB, C5, k=6 & 98.0 & 96.5\\
& RGD, C5, k=6 & 98.1 & 97.2\\
& RGD, C4, k=6 & 97.9 & \bf{97.8}\\
& RGD, C4, k=4 & \bf{98.8} & 97.6\\
\hline
\end{tabular}
\end{center}
\end{table}

\begin{table*}[ht]
\caption{Accuracy Under Different Jaccard Thresholds.}
\label{difjac}
\begin{center}
\begin{tabular}{l|l|l|l|c|c|c|c|c|c|c|c}

\hline
\multirow{3}{*}{\bf{Author}} &  \multirow{3}{*}{{\bf Backbone} }&\multirow{3}{*}{\bf{Anchor Setting}} &\multirow{3}{*}{{\bf Input }}& \multicolumn{8}{c}{\bf{Jaccard Thresholds}} \\
	\cline{5-12} & & & & \multicolumn{4}{c|}{\bf{Image-wise (\%)}} & \multicolumn{4}{c}{\bf{Object-wise (\%)}} \\
	\cline{5-12} & & & & 20\% & 25\%& 30\%& 35\%& 20\%& 25\%& 30\%& 35\% \\
\hline
\multirow{2}{*}{Guo et al.\cite{guo2017hybrid}} & - & 1 scale and 1 aspect ratio& RGB-T & 93.8 & 93.2 & 91.0 & 85.3 & 85.1 & 82.8 & 79.3 & 74.1 \\
& - & 3 scales and 3 aspect ratios & RGB-T& 88.1 & 86.4 & 83.6 & 76.8 & 90.8 & 89.1 & 85.1 & 80.5 \\
Chu et al.\cite{chu2018grasp}& ResNet50 & 3 scales and 3 aspect ratios & RGB-D & - & 96.0 & 94.9 & 92.1 & - & 96.1 & 92.7 & 87.6\\
\hline
\multirow{3}{*}{\bf{Ours} }& VGG-16 & 1 scale and 1 aspect ratio & RGD & 98.0 & 97.5 & 93.8 & 87.5 & 97.0 & 96.4 & 92.9 & 88.1\\
& ResNet-50& 1 scale and 1 aspect ratio & RGD &{\bf98.8} & {\bf98.8} & {\bf96.8} & {\bf94.1} & 97.1 & 97.0 & 94.7 & 90.0 \\
& ResNet-101& 1 scale and 1 aspect ratio & RGD &  {\bf98.8} & {\bf98.8} & 96.4 & 93.7 & {\bf98.1} &{\bf97.6}   & {\bf96.0} & {\bf92.9}\\
\hline

\end{tabular}
\end{center}
\end{table*}

\begin{table*}[ht]
\caption{Accuracy Under Different Angle Thresholds (Jaccard Threshold is 25\%).}
\label{difang}
\begin{center}
\begin{tabular}{l|l|c|c|c|c|c|c|c|c|c|c}

\hline
\multirow{3}{*}{\bf{Backbone}} & \multirow{3}{*}{\bf{Setting}} &  \multicolumn{10}{c}{\bf{Angle Thresholds}} \\
	\cline{3-12} & & \multicolumn{5}{c|}{\bf{Image-wise (\%)}} & \multicolumn{5}{c}{\bf{Object-wise (\%)}} \\
	\cline{3-12} & & $10^\circ$ & $15^\circ$ & $20^\circ$ & $25^\circ$ & $30^\circ$ & $10^\circ$ & $15^\circ$ & $20^\circ$ & $25^\circ$ & $30^\circ$ \\
\hline
 {\bf VGG-16}& \multirow{3}{*}{1 scale and 1 aspect ratio, RGD, C4, k=4}& 79.0 & 90.7 & 95.1 &96.6 & 97.5  & 79.0 & 90.5 & 95.1 & 95.9 & 96.4\\
 {\bf ResNet-50}& & 85.2& {\bf94.0}& {\bf96.7} & {\bf98.1} &  {\bf98.8}& 83.1 & 90.6 & 94.6 & 96.6 & 97.0\\
 {\bf ResNet-101}& &  {\bf86.6} & 93.7 & 96.5 & 97.9 &  {\bf98.8}&  {\bf85.1 }&  {\bf92.1} & {\bf95.4}  & {\bf97.2} &{\bf97.6}\\
\hline

\end{tabular}
\end{center}
\end{table*}

 \begin{figure}[!ht] 
 \center{\includegraphics[scale=0.05]{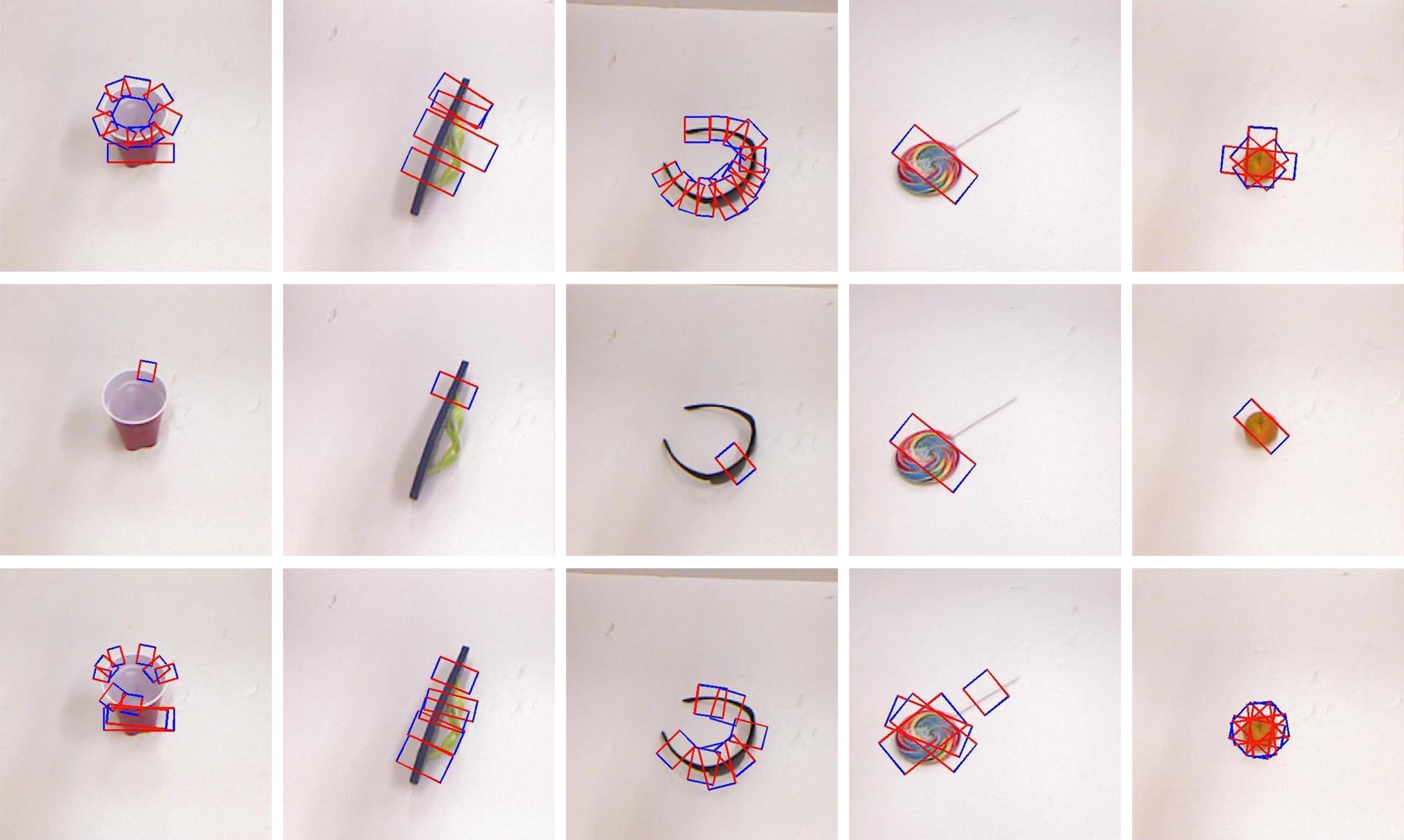}}        
 \caption{True positive examples from ResNet-101 network. The first row is ground-truth oriented rectangles. The second row is the visualization of Top1 grasp detection result. The third row is results of multi grasps.}
 \label{positive}
 \end{figure}
 
 \begin{figure}[t] 
 \center{\includegraphics[scale=0.05]{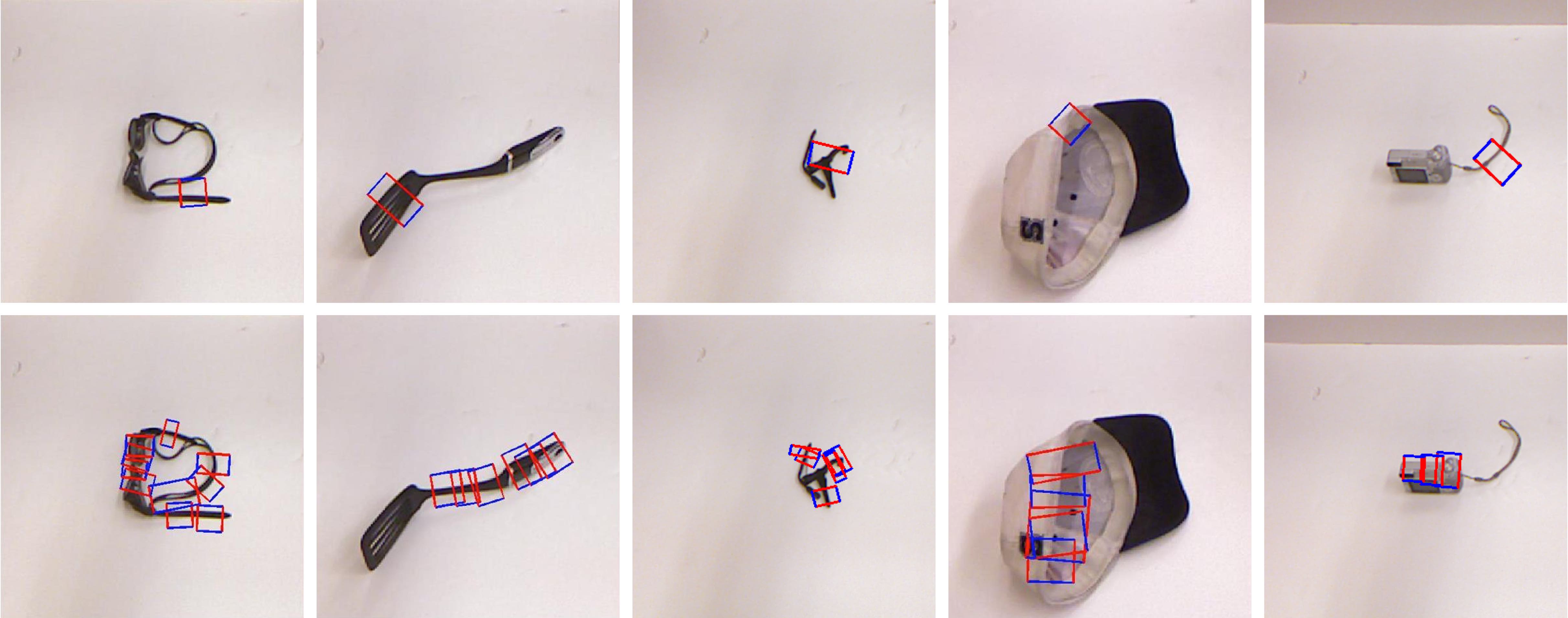}}        
 \caption{Unsuccessful detections from ResNet-101 network. The first row is the detection result. The second row is the ground-truth oriented rectangle. Note that the accuracy of our model cause at most 4 unsuccessful detections at each fold. Therefore, the unsuccessful detections listed here come from different folds.}
 \label{negative}
 \end{figure}
 
 \begin{figure}[ht!]
\center{
\includegraphics[scale=0.05]{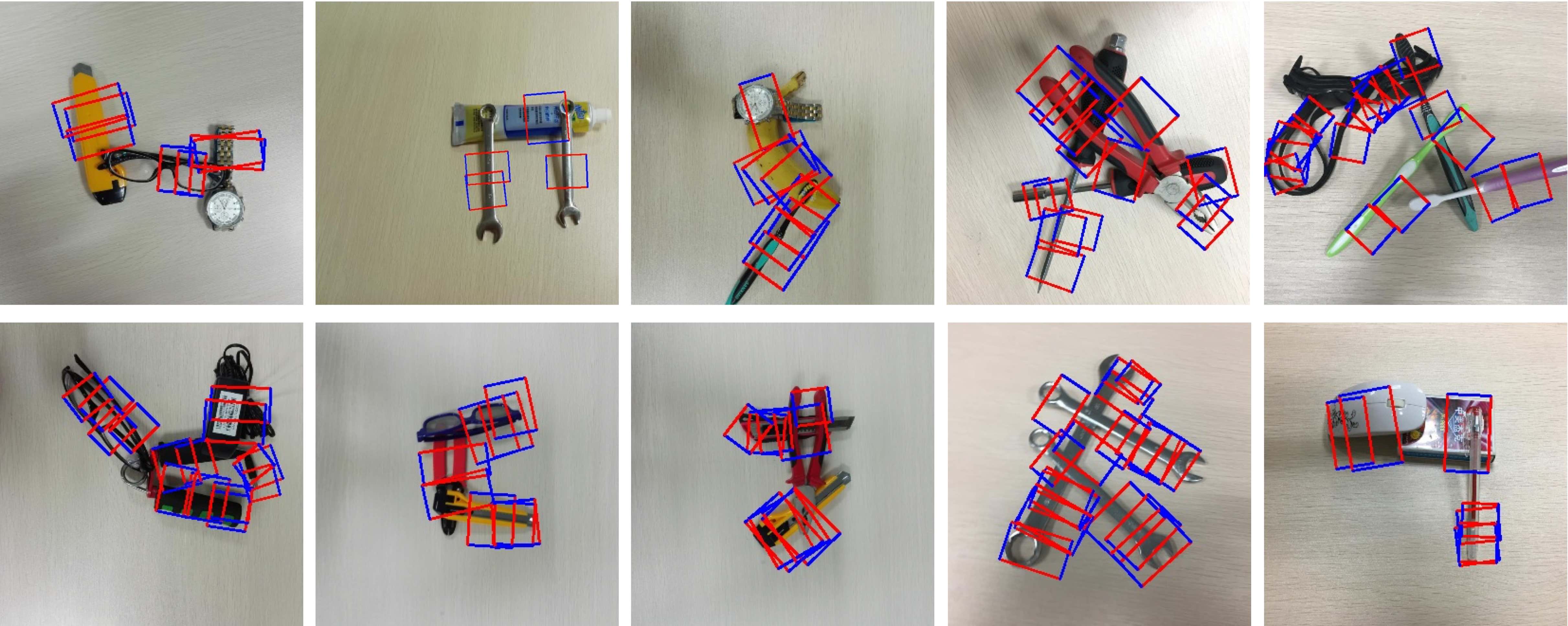}}
\caption{Visualization of detection in more complex scenes and unseen objects.}
\label{multi}
\end{figure}

\begin{table*}[ht]
\caption{Accuracy of Different Methods (Jaccard Threshold is 25\%).}
\label{resultstable}
\begin{center}
\begin{tabular}{l|l|c|c|c}
\hline
\multirow{2}{*}{\bf{Author}} &\multirow{2}{*}{\bf{Algorithm}} &  \multicolumn{2}{c|}{\bf{Accuracy}} & \multirow{2}{*}{\bf{Speed (FPS)}}\\
\cline{3-4} & & \bf{ Image-wise (\%) }&\bf{ Object-wise (\%) }& \\
\hline
Jiang et al.\cite{jiang2011grasp} & Fast Search & 60.5 & 58.3 & 0.02\\
Lenz et al.\cite{lenzgrasp}& SAE, struct. reg. Two stage & 73.9 & 75.6 &  0.07\\
Redmon et al.\cite{redmongrasp}& AlexNet, MultiGrasp &88.0 & 87.1 & 3.31 \\
Kumra et al.\cite{resnetgrasp}& ResNet-50$\times$2, Multi-model& 89.2 & 89.0 & 16.03 \\
Guo et al.\cite{guo2017hybrid}& ZF-net, 3 scales, 3 aspect ratios & 93.2 & 89.1 & - \\
\multirow{2}{*}{Chu et al.\cite{chu2018grasp}}& VGG-16 & 95.5 & 91.7 &17.24 \\
 & ResNet-50 & 96.0 & 96.1 & 8.33\\
\hline
\multirow{3}{*}{\bf{Ours}} & VGG-16 & 98.2 & 96.4 & \bf{118}\\
& ResNet-50 & {\bf98.8} & 97.0 & 105 \\
& ResNet-101 & {\bf98.8} & \bf{97.8} & 67 \\
\hline
\end{tabular}
\end{center}
\end{table*}

For images used to test the network, we just take a center crop without any other augmentation. Then, they are feed into the network one by one, not in the form of mini-batch. We test the proposed models under different Jaccard index thresholds. 

\subsubsection{Self-Comparison Experiments}In Table \ref{rgbrgd}, we demonstrate the influence of different inputs and settings on final results. In the first two experiments of using each backbone, we change the input from RGB to RGD. The performance on image-wise split changes slightly, but there is a considerable improvement in object-wise split. The reason is that when the network uses RGB as input, it tends to detect edges or stripes in the image, but depth information provide a possible way to help network discover the convex points on the object surface. Therefore, RGD input will help our algorithms generalize to new scenes and targets. The rest experiments explore the change of convolutional features and oriented anchor number. We can see that Conv4 features are more suitable for grasp detection because Conv4 provide a finer features including more detailed information of the image than Conv5 with little loss of network depth. Different from object detection, grasps are always represented by small rectangles, which are usually smaller than $80\times80$. Therefore, finer features will be better for grasp detection. Besides, increasing anchor boxes in each grid cell will not have distinct improvement for the final results. Therefore, in our experiments, we use $k=4$.

\subsubsection{Comparison Experiments and Analysis}The detailed results of our methods compared with recent state-of-the-art methods are listed in Table \ref{difjac}. The comparison with the methods of Guo et al.\cite{guo2017hybrid} demonstrates that 3 scales and 3 aspect ratios can largely improve the performance on object-wise split. The reason may be that the appearance of unknown objects requires a thorough detection with multiple rectangle settings. The distribution of scale and aspect ratio for grasp rectangle in training set and test set when using image-wise split is more similar than that when using object-wise split. Therefore, algorithms need to tackle with new position and size, which places more emphasis on the orientation. Although ResNet-101 has similar performance with ResNet-50 on image-wise split, it performs better than ResNet-50 on object-wise split, which may be caused by the more powerful feature extraction ability. Using ResNet-101 as feature extractor can be more robust to the change of objects. The performance difference between ResNet-50 and ResNet-101 shows that employing deeper network is another way to improve object-wise accuracy apart from multiple scales and aspect ratios. Our models based on VGG16, ResNet-50 and ResNet101 run at the speed of 118 FPS, 105 FPS and 67 FPS, respectively.

We have emphasized the importance of rotation angles in former sections. In Table \ref{difang}, our models are tested under different angle thresholds. The performance of our model is still good when the threshold is bigger than $15^{\circ}$. Note that the deeper network brings little improvement on the accuracy of image-wise split. In practice, we have $600$ $(10\times10\times k, k = 6)$ predictions, of which the number is only one third of that in Chu et al.\cite{chu2018grasp}. In other words, we achieve better performance using a less dense prediction. Our ResNet-50 model outperforms the state-of-the-art method of Chu et al. by 2.6\% and 0.9\% on image-wise split and object-wise split, respectively, which demonstrates that oriented anchor box mechanism provides a more accurate and efficient way for grasp detection. In view of the great improvement brought by multiple scales and aspect ratios on object-wise split accuracy, further improvement on our accuracy of object-wise split can be achieved by introducing multiple scales.

Table \ref{resultstable} shows the comparison with other algorithms on Cornell Dataset. We can see that our algorithms achieve the highest accuracy and fastest speed. Our model fwith ResNet-50 backbone balances performance and speed best with 105 FPS and 97.0\% accuracy on object-wise split. The best results are obtained by our ResNet-101 model with 97.8\% accuracy on object-wise split. Though the speed of our ResNet-101 model is only 2/3 of ResNet-50 model, it is nearly 4 times faster than the previous fastest method (VGG-16 of Chu et al.\cite{chu2018grasp}). All our models can meet the real-time requirement in real-world robot task.

\subsubsection{Results Display}In Fig.\ref{positive}, we visualize ground-truth rectangles and detection results of some objects in the test set of Cornell Grasp Dataset under image-wise splitting. The first row shows the ground-truth grasp rectangles of the objects. The second row visualizes the Top1 detection results of these objects. Multi-grasp detection results are demonstrated in the third row. All the grasp rectangles in the third row have a graspable score over 0.5. The multi-grasp result of first row reveals the working mechanism of oriented anchor box, which predicts several rectangles at the same center but with different rotation angles. From the multi-grasp examples of the second line, we can see that our model predicts grasps from the shape of the objects rather than just fitting the annotation. Note that the grasp rectangles in multi-grasp detection results cover most graspable positions, even the one not labeled in ground-truth. Only a few predicted grasps have large overlap with others, which indicates our predicted grasps efficiently cover most of the representative grasps.
 
Some false detection results are shown in Fig.\ref{negative}. The first row is the detection results and the second row is the ground-truth rectangles. Although these predictions do not satisfy the rectangle metric, they are still feasible because the annotation is not exhaustive.

To test our model on more realistic and complex scenes, we test our ResNet-101 model, which is trained on image-wise split, with some pictures where the objects overlap with each other. Results are shown in Fig.\ref{multi}, Some categories (knife, watch, wrench, wrist developer) never appear in Cornell Grasp Dataset. Despite the occlusion, our model still has good performance under more realistic and complex scene. It indicates that our models have satisfying generalization ability.

\section{Robotic Experiments}

\subsection{Method}

During robotic experiment, there are several problems from grasp detection to robotic grasp:
 
\begin{itemize}
\item 1) Detected grasps are 5 dimension vectors in image coordinate system. Therefore, how can the robot locate the best grasp position in real world?
\item 2) How does the robot decide the best grasp posture?
\item 3) How does the robot plan the trajectory to finish grasp?
\end{itemize}

 \begin{figure}[!t] 
 \center{\includegraphics[scale=0.13]{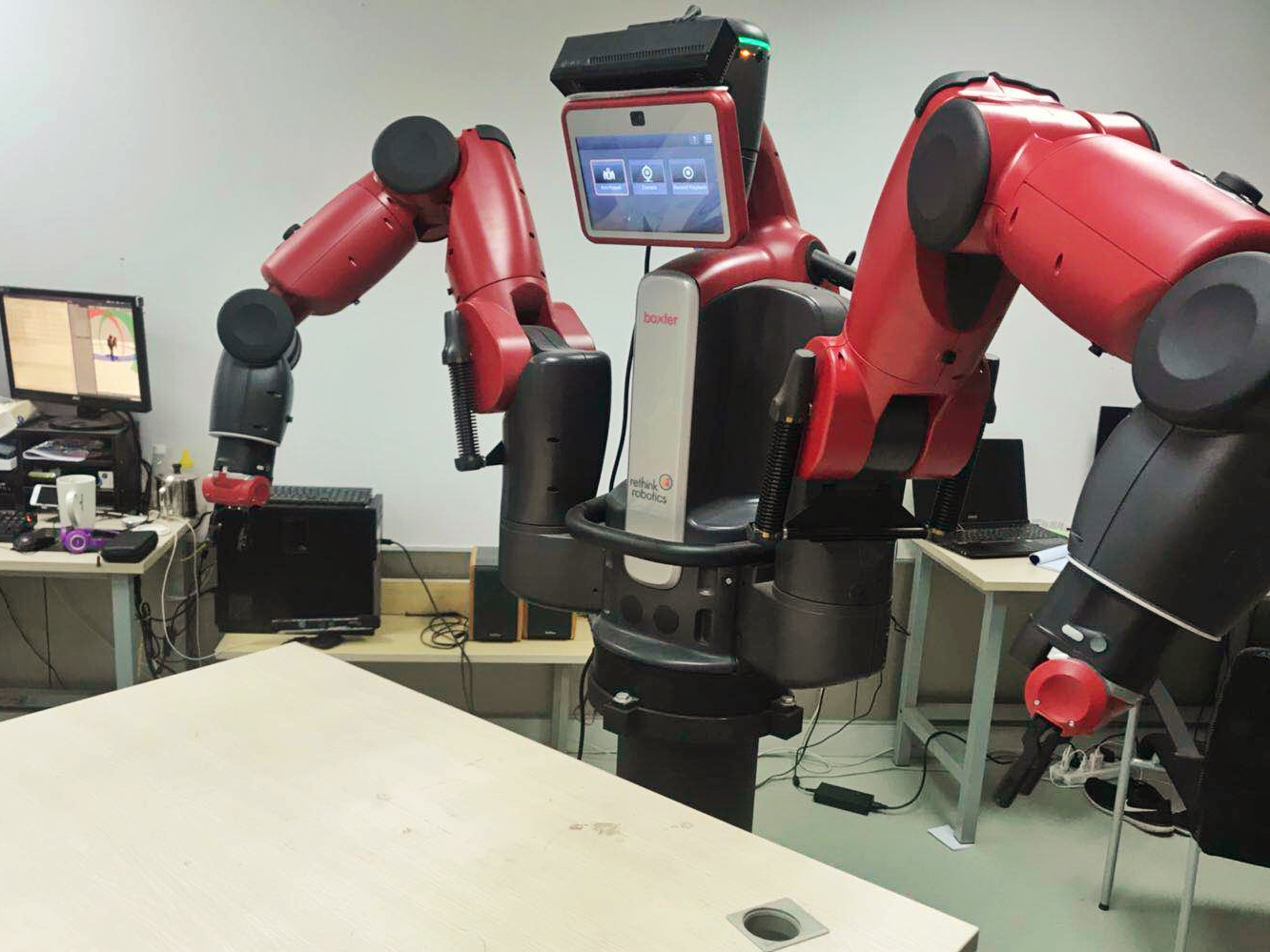}}        
 \caption{Environment of robotic experiment.}
 \label{expenv}
 \end{figure}

 \begin{table*}[h]
\caption{Success Rate of Robotic Experiments on Different Sorts of Objects}
\label{roboexp}
\begin{center}
\begin{tabular}{p{2.5cm}|p{1.5cm}<{\centering}|p{1.5cm}<{\centering}||p{2.5cm}|p{1.5cm}<{\centering}|p{1.5cm}<{\centering}}
\hline
\hline
\multicolumn{3}{c||}{\bf{Familiar Things}}&\multicolumn{3}{c}{\bf{Unseen Things}}\\
\hline
Object & Suc./Tr. & Det./Tr.  &Object & Suc./Tr. & Det./Tr.  \\
\hline
\multicolumn{6}{c}{\bf{Fruits}} \\
\hline
Banana & 8 / 10 & 10 / 10  & Peach & 8 / 10 & 10 / 10 \\
Apple & 9 / 10 & 10 / 10 & Peer & 8 / 10 &10 / 10 \\
Orange & 9 / 10 & 10 / 10 & Avocado & 9 / 10 & 10 / 10 \\
\hline
Average(\%) & 86.7\% & 100\% & Average(\%) & 83.3\% & 100\% \\
\hline
\multicolumn{6}{c}{\bf{Tools}} \\
\hline
Scissors & 9 / 10 & 10 / 10 & Wrench & 10 / 10 & 10 / 10  \\
Screwdriver & 10 / 10 & 10 / 10 & Pliers & 9 / 10 & 10 / 10 \\
Tape & 7 /10 & 8 / 10 & Electroprobe & 10 / 10 & 10 / 10 \\
\hline
Average(\%) & 86.7\% & 93.3\% & Average(\%) & 96.7\% & 100\% \\
\hline
\multicolumn{6}{c}{\bf{Containers}} \\
\hline
Colorful Box & 10 / 10 & 10 / 10  & Ice Cube Tray & 7 / 10 & 10 / 10  \\
Cup & 8 / 10 & 10 / 10 & Spectacle Case & 10 / 10 & 10 / 10 \\
Bowl & 10 / 10 & 10 / 10  & Glass* & 5 / 10 & 6 / 10 \\
\hline
Average(\%) & 93.3\% & 100\% & Average(\%) & 73.3\% & 86.7\% \\
\hline
\multicolumn{6}{c}{\bf{Others}} \\
\hline
Glasses & 9 / 10 & 10 / 10 & Wrist Developer & 10 / 10 & 10 / 10  \\
Stapler & 9 / 10 & 10 / 10 & Push Up Bar & 7 / 10 & 10 / 10 \\
Spoon & 10 / 10 & 10 / 10 & Table Tennis Bat & 8 / 10 & 8 / 10 \\
\hline
Average(\%) & 93.3\% & 100\% & Average(\%) & 83.3\% & 93.3\% \\
\hline
\hline
Total(\%) & 90.0\% & 98.3\% & Total(\%) & 84.2\% & 95.0\% \\
\hline
\multicolumn{6}{l}{* means that the experiment is done using RGB model.}

\end{tabular}
\end{center}
\end{table*}

 \begin{figure*} [!t]
 \center{\includegraphics[scale=0.08]{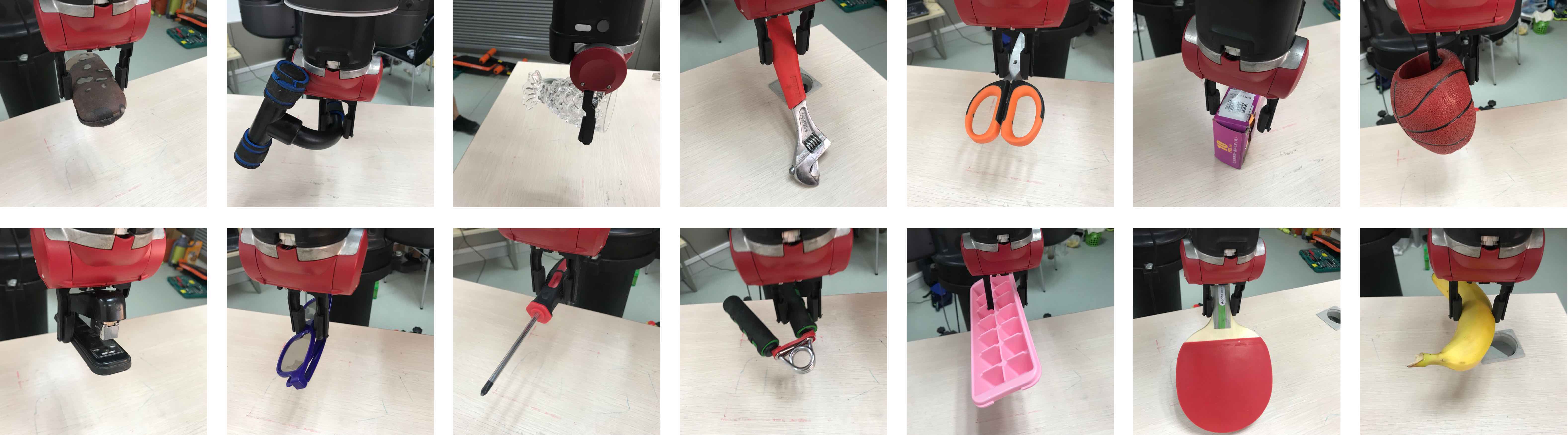}}        
 \caption{Robot experiments.}
 \label{robexp}
 \end{figure*}

Following Lenz et al.\cite{lenzgrasp}, we solve the last two problems directly: using the average surface normals around the best grasp position as the grasp vector, along which the gripper approaches the target, and using robot inverse kinematics package to plan grasp trajectories. However, different from Lenz et al.\cite{lenzgrasp}, we locate the best grasp position following these steps: 1) use all grasps of which the graspable confidence scores are more than 0.5 as grasp candidates; 2) choose the closest candidate away from the center of target as the grasp rectangle to be executed; 3) use the point with minimum depth in the chosen grasp rectangle as the grasp point, around which the grasp vector is estimated; 4) map the grasp point and grasp vector in image coordinate system to robot coordinate system using camera-robot calibration. Because we do not detect target during grasp, we estimated the object center $(x_{oc},y_{oc})$ using following Eq.\ref{estcenter}:

\begin{equation}
\label{estcenter}
\begin{split}
x_{oc} = \frac{max(\{x\}_{candi}) + min(\{x\}_{candi})}{ 2}\\
y_{oc} = \frac{max(\{y\}_{candi}) + min(\{y\}_{candi})}{ 2}\end{split}
\end{equation}

In Eq.\ref{estcenter}, $\{x\}_{candi}$ and $\{y\}_{candi}$ is the set of $x$ and $y$ coordinates of predicted candidate centers.
 
\subsection{Environment Setting}
 
The robot used in our experiments is Baxter robot designed by RethinkRobotics Corporation. Baxter robot has 2 arms and each arm has 7 DoF. Baxter robot is fixed on the same height of the desktop. RGB-D camera used in our work is Kinect V2 developed by Microsoft and it is fixed on the head of the baxter. The baxter gripper has two parallel fingers, but the active range is limited (about 4 cm). Therefore, before each experiment, we need to manually adjust the width of the gripper to fit the specified target size. Complete environment is shown in Fig.\ref{expenv}.

To detect grasp, the Kinect V2 has a $60^{\circ}$ depression angle relative to the desktop, and limited by the robot arm and inverse kinematics, targets should be put into a predefined area which is around $30cm\times30cm$ on the desk. The calibration between robot and camera is done using 4 calibration points in real-world. To ensure the precise calibration, we recommend these 4 points should be taken with the following coordinates in real-world: $(0,0,0)$, $(L,0,0)$, $(0,L,0)$ and $(L,L,L)$ as Lenz et al. do, where $L$ is chosen by yourself. In our experiment, $L$ is about 15 cm. After selecting the calibration points, through the their coordinates in robot coordinate system and image coordinate system, we can estimate the transformation matrix, which can map any point coordinates from one to the other one.

To evaluate the performance of our proposed algorithm more validly, we use objects from 4 sorts including familiar (belonging to the category that has appeared in Cornel Dataset) and unknown things: fruits, tools, containers and others. In each sort, there are totally 5 objects. With each object, we try some times to grasp it and record the number of success.

\subsection{Results}

Robotic experimental results are shown in Table \ref{roboexp}. In robotic experiments, we use RGD model based on ResNet101 backbone to detect grasps. We can see that our algorithm achieves a 90.0\% and 84.2\% success rate on familiar things and unseen things respectively in real-world robotic grasp. Some pictures of grasps are listed in Fig. \ref{robexp}. Note that all things used in robotic experiment are not seen during training. Results show that our algorithm can tackle different things well, and will help the robot to successfully grasp the target.

Though detection results are satisfactory, the success rate of robotic grasp is not as good as detection results. It's worth noting that we come across some objects that will cause a low success rate in robotic experiments, such as glass, push up bar and tape. The reasons are summarized as follow:

\begin{itemize}
\item Because of the limitation of Kinect Camera, of which the depth sensor will be invalid when the target is transparent ($e.g.$ a glass), RGD models will not work well with transparent things such as a glass.
\item Objects that have similar color with the background will increase the difficulty of grasp detection.
\item Objects with complex shape ($e.g.$ a push up bar) may hinder the gripper from successful grasping due to the limitation of the Baxter gripper ($e.g.$ limited length and active range).
\item Flat things with circular surface such as a table tennis bat may occasionally confuse the network and be treated as a plate and the detected grasps may be located on the margin and cause failure grasping.
\end{itemize}

Therefore, in our experiment, when grasping a glass, we use the RGB model instead of the RGD model. However, results are not satisfying as well. This is caused by the similar color with the background and loss of depth information. The other failures are mostly caused by the error of reaching action of Baxter. In the future, we will try visual servo methods to improve the performance of the Baxter robot.

\section{Conclusion}
In this paper, we build a robust and real-time robotic grasp approach based on fully convolutional neural network. We propose Oriented Anchor Box Mechanism to fit the property of robotic grasps, and it is proved to be efficient and helpful to improve the performance of grasp detection. Besides, Angle Matching is proposed with Oriented Anchor Box to accelerate the convergence of our network. Experiments show that our method achieves an accuracy of 98.8\% and 97.8\% in image-wise split and object-wise split respectively, and the speed of our detection algorithm is up to 67 FPS with GTX 1080Ti, which means our method outperforms all the current state-of-the-art grasp detection algorithms on Cornell Dataset both in speed and accuracy. Besides, it is proved that our algorithm can generalize well in complex scenes and unseen objects.

In robotic grasp experiment, we use grasp detection result to find the grasp point and grasp vector. Using coordinate calibration, we map the grasp vector from camera frame to Baxter frame, and make Baxter robot execute the grasp configuration using inverse kinematics. Our robotic grasp approach achieves an average success rate of 90.0\% and 84.2\% of familiar things and unseen things respectively on Baxter robot platform..


%


\section*{Acknowledgment}

This work was supported in part by the key project of Trico-Robot plan of NSFC under grant No. 91748208, National Key Program of China No.2017YFB1302200,key project of Shaanxi province No.2018ZDCXL-GY-06-07, and NSFC No.61573268.

\ifCLASSOPTIONcaptionsoff
  \newpage
\fi



\bibliographystyle{IEEEtran}
\bibliography{IEEEabrv}
%



%

\begin{IEEEbiography}[{\includegraphics[width=1in,height=1.25in,clip,keepaspectratio]{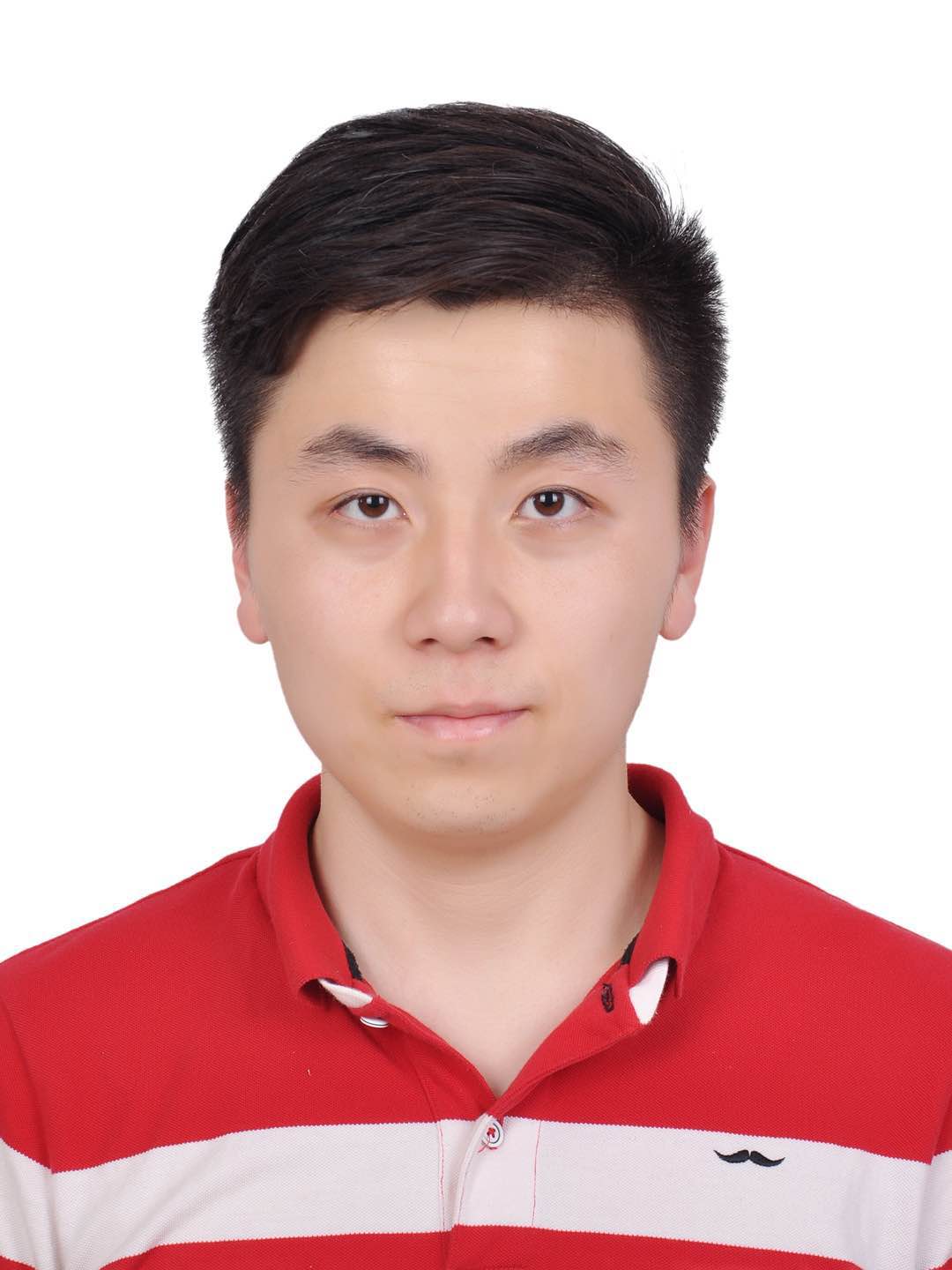}}]{Hanbo Zhang}
received the B.S. degree in
Information Engineering from Xi'an Jiaotong University, Xi'an,
China, in 2016, where he is currently pursuing the
Ph.D. degree with the School of Electronic and
Information Engineering.
His current research interests include robotic grasp,
policy search in robotics and visual relationship detection.
\end{IEEEbiography}

\begin{IEEEbiography}[{\includegraphics[width=1in,height=1.25in,clip,keepaspectratio]{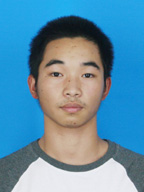}}]{Xinwen Zhou}
received the B.S. degree in
automation from Xi'an Jiaotong University, Xi'an,
China, in 2016, where he is currently pursuing the
Master degree with the School of Electronic and
Information Engineering.
His current research interests include robotic grasp and object detection.
\end{IEEEbiography}

\begin{IEEEbiography}[{\includegraphics[width=1in,height=1.25in,clip,keepaspectratio]{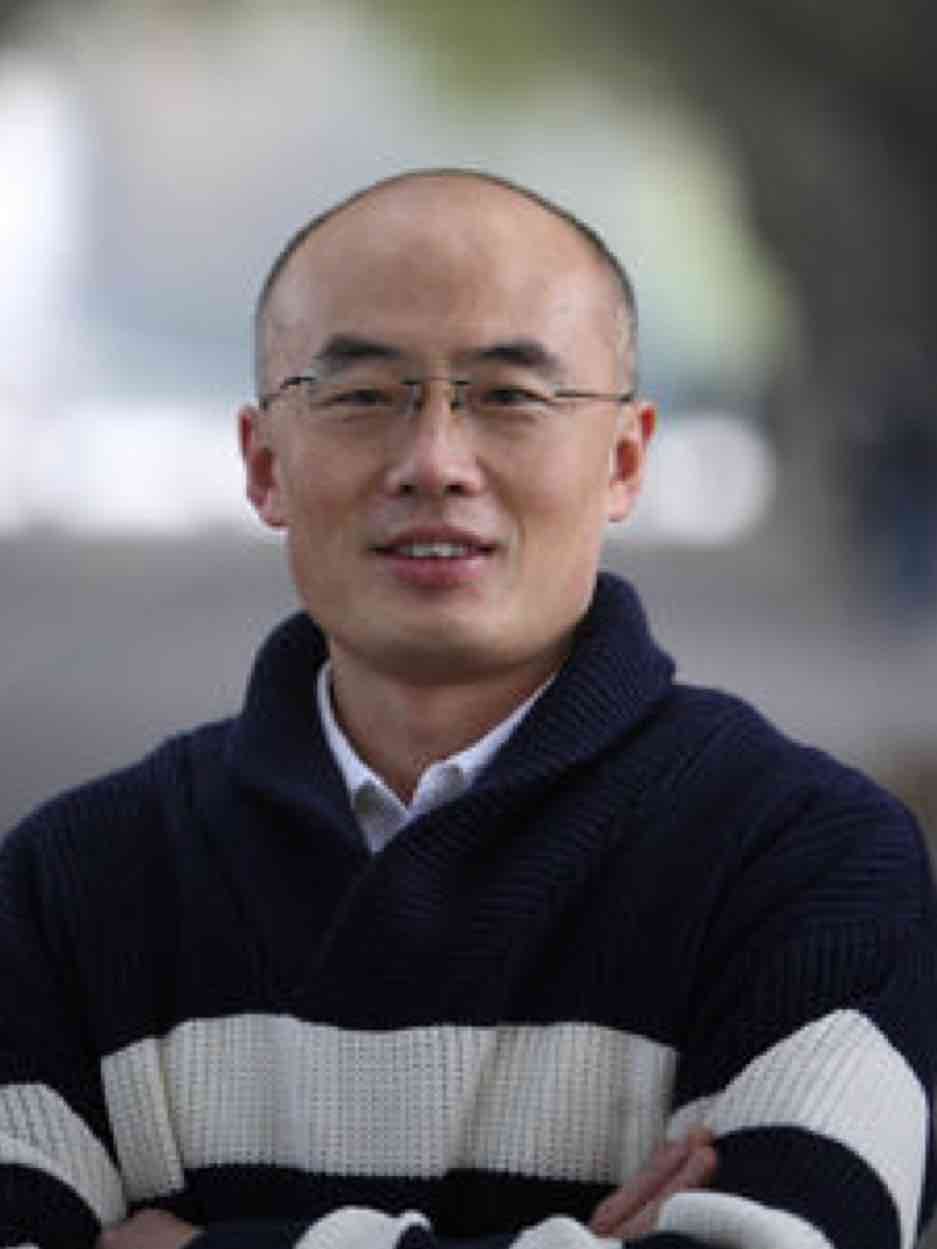}}]{Xuguang Lan}
 (M'06) received Ph.D. degree in Pattern Recognition and Intelligent System from Xi'an Jiaotong University in 2005. He was a postdoctor in department of computer science, Xi'an Jiaotong University from Dec. 2005 to Jan. 2008. He was a visiting scholar of Ecole Centrale de Lyon from May.2005 to Oct.2005, and Northwestern University from Sep. 2013 to Oct. 2014. Currently, he is a professor at Institute of Artificial Intelligence and Robotics in Xi'an Jiaotong University. His research interests include computer vision, machine learning, pattern recognition, human-robot collaboration, and content-based image/video coding. He is a member of IEEE.
\end{IEEEbiography}

\begin{IEEEbiography}[{\includegraphics[width=1in,height=1.25in,clip,keepaspectratio]{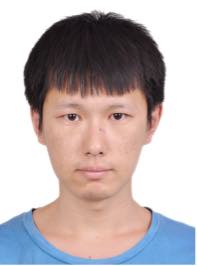}}]{Jin Li}
received the B.S. degree in automation from
Xi'an Jiaotong University, Xi'an, China, in 2013, and
is currently working toward the Ph.D. degree at the
School of Electronic and Information Engineering,
Xi'an Jiaotong University.
His research interests including quantization technique
for approximate nearest neighbor search and
large-scale retrieval.
\end{IEEEbiography}

\begin{IEEEbiography}[{\includegraphics[width=1in,height=1.25in,clip,keepaspectratio]{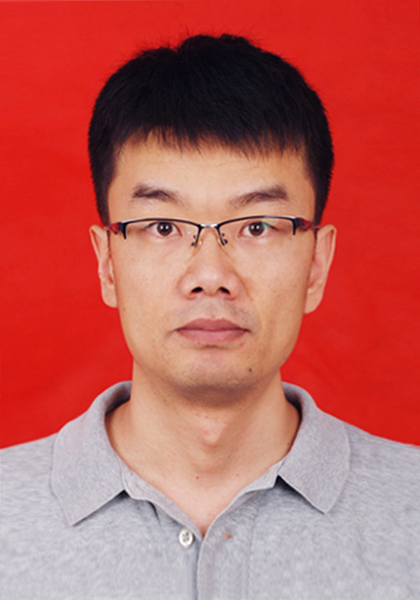}}]{Zhiqiang Tian}
received the B.S. degree in Automation Control from the Northeastern University, in 2004, and received the M.S. and Ph.D. degrees in Control Science and Engineering from Xi'an Jiaotong University, in 2007 and 2013, respectively. Now he is an associate professor at the School of Software Engineering, Xi'an Jiaotong University. His research interests are image/video processing, computer vision, medical image analysis, and robotics.
\end{IEEEbiography}


\begin{IEEEbiography}[{\includegraphics[width=1in,height=1.25in,clip,keepaspectratio]{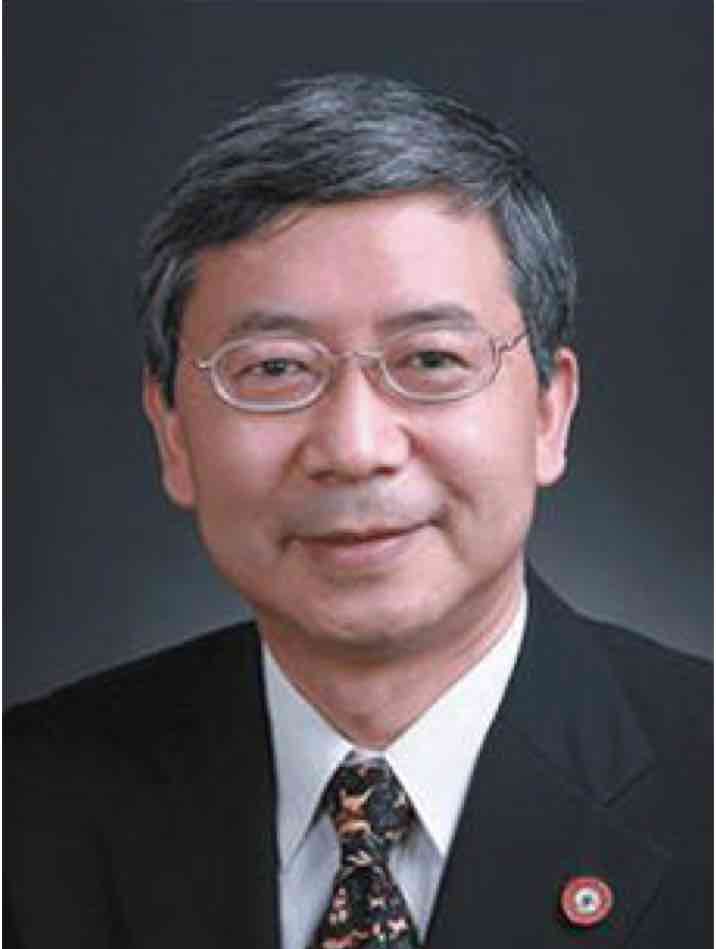}}]{Nanning Zheng}
 (SM'93-F'06) reeived the B.S.
and M.S. degrees in information and control engineering
from Xi'an Jiaotong University (XJTU),
Xi'an, China, in 1975 and 1981, respectively, and
the Ph.D. degree in electrical engineering from Keio
University, Yokohama, Japan, in 1985.
In 1975, he joined XJTU, where he is currently
a Professor and the Director of the Institute
of Artificial Intelligence and Robotics. His current
research interests include computer vision, pattern
recognition and image processing, and hardware
implementation of intelligent systems.
Dr. Zheng became a member of the Chinese Academy of Engineering,
in 1999, and the Chinese Representative on the Governing Board of the
International Association for Pattern Recognition. He also serves as the
Executive Deputy Editor for the Chinese Science Bulletin and an Associate
Editor for the IEEE TRANSACTIONS ON INTELLIGENT TRANSPORTATION
SYSTEMS.
\end{IEEEbiography}




\end{document}